\documentclass[conference]{IEEEtran}
\IEEEoverridecommandlockouts
\usepackage[noadjust]{cite}
\usepackage{amsmath}
\usepackage{amssymb}
\usepackage{xspace}
\usepackage{xcolor}
\usepackage{graphicx}

\usepackage[pagebackref,breaklinks,colorlinks]{hyperref}
\usepackage[capitalize]{cleveref}
\crefname{section}{Sec.}{Secs.}
\Crefname{section}{Section}{Sections}
\Crefname{table}{Table}{Tables}
\crefname{table}{Tab.}{Tabs.}

\usepackage{booktabs}
\usepackage{algorithm}
\usepackage{algpseudocode}
\usepackage{subcaption}
\usepackage{comment}

\DeclareMathOperator*{\argmax}{arg\,max}

\newcommand{\set}[1]{\mathcal{#1}}

\makeatletter
\DeclareRobustCommand\onedot{\futurelet\@let@token\@onedot}
\def\@onedot{\ifx\@let@token.\else.\null\fi\xspace}

\def\eg{\emph{e.g}\onedot} 
\def\ie{\emph{i.e}\onedot} 
 
\def\etc{\emph{etc}\onedot} \def\vs{\emph{vs}\onedot}
 
\def\etal{\emph{et al}\onedot}
\makeatother

\def\BibTeX{{\rm B\kern-.05em{\sc i\kern-.025em b}\kern-.08em
    T\kern-.1667em\lower.7ex\hbox{E}\kern-.125emX}}
    
\begin{document}

\title{Exploring Active Learning for Label-Efficient Training of Semantic Neural Radiance Field}

\author{Yuzhe Zhu$^{*\dag2}$ \quad Lile Cai$^{*1}$ \quad Kangkang Lu$^1$ \quad Fayao Liu$^1$ \quad Xulei Yang$^1$\\
$^1$Institute for Infocomm Research (I\textsuperscript{2}R), A*STAR, Singapore\\
$^2$Nanyang Technological University, Singapore\\
\texttt{s230061@e.ntu.edu.sg,\{caill,lu\_kangkang,liu\_fayao,yang\_xulei\}@i2r.a-star.edu.sg} \\

\thanks{* Equal contribution.}
\thanks{$\dag$ Work done during internship with I\textsuperscript{2}R.}
}
\maketitle

\begin{abstract}
Neural Radiance Field (NeRF) models are implicit neural scene representation methods that offer unprecedented capabilities in novel view synthesis. Semantically-aware NeRFs not only capture the shape and radiance of a scene, but also encode semantic information of the scene. The training of semantically-aware NeRFs typically requires pixel-level class labels, which can be prohibitively expensive to collect. In this work, we explore active learning as a potential solution to alleviate the annotation burden. We investigate various design choices for active learning of semantically-aware NeRF, including selection granularity and selection strategies. We further propose a novel active learning strategy that takes into account 3D geometric constraints in sample selection. Our experiments demonstrate that active learning can effectively reduce the annotation cost of training semantically-aware NeRF, achieving more than 2$\times$ reduction in annotation cost compared to random sampling.
\end{abstract}

\begin{IEEEkeywords}
Active Learning, Semantic Neural Radiance Field 
\end{IEEEkeywords}

\section{Introduction}
\label{sec:intro}
Neural Radiance Field (NeRF) models \cite{mildenhall2021nerf} have recently emerged as a powerful tool for 3D scene representation. It represents the geometry and radiance of a single scene with a neural network and performs novel view synthesis via volume rendering. NeRF models have found a wide range of applications in augmented reality, autonomous navigation, urban mapping, and more \cite{gao2022nerf}.

Traditional NeRFs primarily focus on geometric and photometric accuracy \cite{zhang2020nerf++,yu2021pixelnerf}. Semantic-NeRF \cite{zhi2021place} marks a significant advancement by jointly representing the physical characteristics and semantics of a scene. It adds a semantic prediction branch that maps spatial coordinates to semantic labels. This leap in technology facilitates more sophisticated applications such as scene understanding and editing.

Unlike geometry and radiance that can be trained using only (unlabelled) multi-view images, semantics are human-defined concept and some form of labelling would always be needed. In \cite{zhi2021place}, Semantic-NeRF has been shown to achieve remarkable performance with sparse annotation. However, only image-level random sampling is investigated for sparse labelling. It is not clear how the performance can be further boosted by employing more sophisticated sampling techniques. 

In this work, we explore active learning (AL) as a promising solution to alleviate the annotation cost for training Semantic-NeRF. Active learning has been extensively studied in various visual tasks, including image classification, semantic segmentation and object detection \cite{ren2021survey}, but it has not been investigated for the newly emerging semantically-aware NeRF models. The most close work is ViewAL \cite{siddiqui2020viewal}, which exploits viewpoint consistency in multi-view datasets for active learning of semantic segmentation models. However, different from frame-level segmentation models (\eg, DeepLabv3+ \cite{chen2018encoder} used in \cite{siddiqui2020viewal}), the semantic prediction branch of NeRF is by construction multi-view consistent (since it is modeled as a viewpoint-invariant function), making viewpoint consistency ineffective for active learning of Semantic-NeRF. In this work, we propose a novel active learning strategy that takes into account 3D geometric constraints in sample selection for Semantic-NeRF.

\begin{figure}[ht]
\vspace{-0.4cm}
\centering
\includegraphics[width=0.7\linewidth]{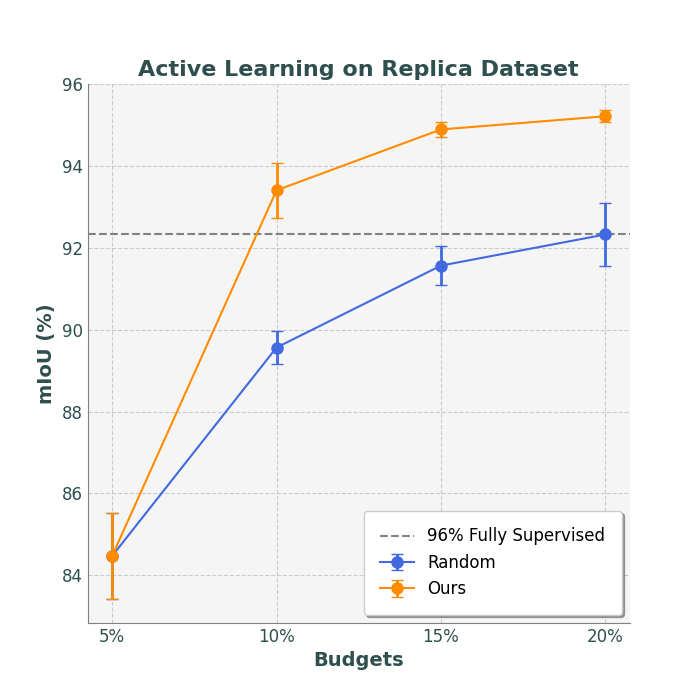}
\caption{Our work demonstrates that active learning can significantly outperform random sampling and serves as a promising solution for label-efficient training of semantically-aware NeRF. }
\label{fig:teaser}
\end{figure}

Our contributions can be summarized as below:

\begin{itemize}
    \item We perform a comprehensive study on active learning for semantically-aware NeRF. We investigate various design choices including selection strategies (\eg, uncertainty-based, diversity-based and hybrid methods) and selection granularity (\eg, image-level \vs region-level selection). Our experiments demonstrate that active learning can effectively reduce the annotation cost for training semantically-aware NeRF, achieving more than 2$\times$ reduction in annotation cost compared to random sampling (\cref{fig:teaser}).
    \item We propose a novel active learning strategy that takes into account 3D geometric constraint in sample selection for semantically-aware NeRF. We incorporate the geometric constraint into the result diversification framework and solve it efficiently using a 2-approximation greedy algorithm.
\end{itemize}

\section{Related Work}
\label{sec:related_work}
\subsection{Label Efficient Learning of Semantically-Aware NeRF}
Neural Radiance Field (NeRF) models \cite{mildenhall2021nerf} have recently emerged as a powerful tool for novel view synthesis. Traditional NeRFs primarily focus on geometric and photometric accuracy \cite{zhang2020nerf++,yu2021pixelnerf}. Semantic-NeRF \cite{zhi2021place} is a groundbreaking work that adds semantic class prediction to density and color prediction. Experiments in \cite{zhi2021place} demonstrate its capability to achieve remarkable performance with sparse labelling. However, only random sampling is investigated in \cite{zhi2021place}, while in this work we conduct a more comprehensive study by investigating various sampling strategies. Liu \etal \cite{liu2023unsupervised} proposed to train a Semantic-NeRF \cite{zhi2021place} model for each scene in a self-supervised fashion by utilizing the pseudo labels produced by a separate frame-level semantic network. Panoptic NeRF \cite{fu2022panoptic} performs joint geometry and semantic optimization by using both 3D and 2D weak semantic information. Liu \etal \cite{liu2023weakly} enabled open-vocabulary segmentation with NeRF by exploiting pre-trained foundation models in a weakly supervised manner, where text descriptions of the objects in a scene are used as weak labels to guide the class assignment. Interactive segmentation of radiance fields has also been investigated \cite{zhi2021ilabel,tang2023scene}, where users are required to manually select which samples to label on 2D views. Instead of relying on users to select queries, our work develops active learning strategies to automatically select the most informative samples to label.

Previous works on label efficient learning of semantically-aware NeRF focus on utilizing pseudo labels generated by separate models to supervise the training of NeRF. However, pseudo labels are not guaranteed to be correct and ground truth labels are still imperative to achieve performance close to fully-supervised learning. In this work, we explore active learning as an alternative solution for label efficient learning of semantic NeRF.

\subsection{Active Learning for Visual Tasks}
As a promising technique to alleviate the annotation burden for training deep models, active learning has been extensively studied for a wide range of tasks, including image classification \cite{beluch2018power,sener2017active,ash2019deep,he2024hybrid,wen2024active}, semantic segmentation \cite{casanova2020reinforced,cai2021revisiting,cai2021exploring}, object detection \cite{wu2022entropy,lyu2023box,liao2024box,zhang2024employing}, and more \cite{ren2021survey}. Depending on the criterion used to select samples, various methods can be grouped into three categories, including uncertainty-based \cite{beluch2018power,yoo2019learning,liu2021influence}, diversity-based \cite{sener2017active} and hybrid methods \cite{ash2019deep,he2024hybrid,wu2021hal}. Uncertainty-based methods select samples that the model is most uncertain about, where uncertainty can be measured by entropy \cite{shannon1948mathematical}, model ensembles \cite{beluch2018power}, learned loss \cite{yoo2019learning}, influence function \cite{liu2021influence}, \etc. Diversity-based methods aim to select a subset of samples that well represent the training data distribution. Sener and Savarese \cite{sener2017active} proposed to select samples that minimize the core-set loss. Hybrid methods consider both uncertainty and diversity in selection. BADGE \cite{ash2019deep} performs selection by applying K-Means++ seeding algorithm on gradient embeddings. UWE \cite{he2024hybrid} generalizes the gradient embedding of BADGE as uncertainty-weighted embeddings, which can be used with arbitrary loss functions and be computed more efficiently.

Active learning for NeRF models is much less explored. ActiveNeRF \cite{pan2022activenerf} investigated active learning for the default (non-semantic) NeRF by selecting views that bring the most reduction in uncertainty for training. To the best of our knowledge, active learning for semantic NeRF models has not been explored in the literature. The work most related to ours is ViewAL \cite{siddiqui2020viewal}, which is developed for 2D semantic segmentation task that exploits model prediction consistency across viewpoints in multi-view datasets. However, as NeRF is constrained to be multi-view consistent by restricting the semantics and density prediction to be independent of viewing direction, the multi-view consistency criterion advocated by ViewAL is ineffective for active learning of semantic NeRF models. In this work, we propose to employ 3D spatial diversity as a more effective selection criterion for NeRF models.

\section{Method}
\label{sec:method}
\begin{figure}[ht]
\centering
\includegraphics[width=0.95\linewidth]{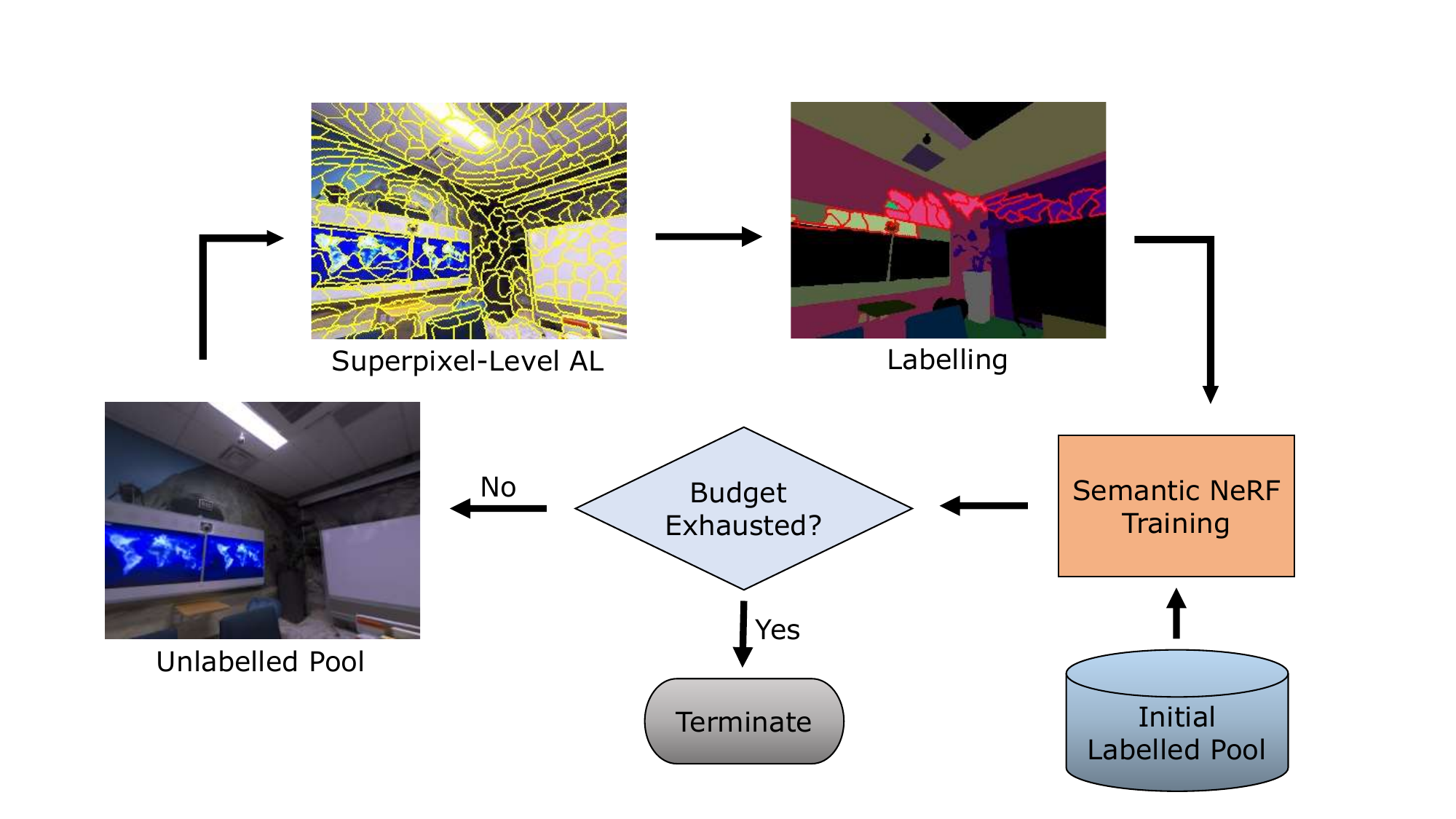}
\caption{The system diagram of the proposed active learning method. A Semantic-NeRF model is first trained on an initially labelled pool. The trained model is then used to evaluate the uncertainty and diversity of unlabelled samples and a batch of most informative samples are selected. We perform selection and annotation at superpixel level, which is shown to be more cost-effective than image level approach. The Semantic-NeRF model is then retrained with all the labelled samples and the process iterates until the annotation budget is exhausted.}
\label{fig:system_diagram}
\end{figure}

The system diagram of our active learning method for training Semantic-NeRF is presented in \cref{fig:system_diagram}. Active learning is an iterative process, where at each iteration, a batch of unlabelled samples are selected for labelling using some selection criterion. The model is then retrained with all the samples labelled so far. The process iterates until the annotation budget is exhausted or the target model performance is met. In the following, we first present the formulation of our active learning method, followed by detailed description of the proposed active selection strategy with 3D geometric constraint. 

\subsection{Problem Formulation}

Our method is inspired by the diversification problem in search engine. When the search engine returns results for a user query, there is a trade-off between having more relevant results and having more diverse results in the top positions. This trade-off between relevancy and diversity mimics the bi-criteria selection strategy of hybrid AL, \ie, the batch of selected samples are desired to be both uncertain and diverse. This motivates us to formulate the AL selection as a diversification problem, which can be solved efficiently by a 2-approximation greedy algorithm \cite{gollapudi2009axiomatic}.

Specifically, letting \(\set{U}_t\) denote the set of unlabelled samples at iteration $t$, the objective is to find a set \(\set{B}_t\) that satisfies the following constraints:
\begin{equation}
\mathcal{B}_t^* = \argmax_{\mathcal{B}_t \in U_t, |\mathcal{B}_t|=K} f(\mathcal{B}_t, u(\cdot), d(\cdot, \cdot)),
\label{eq:obj}
\end{equation}
where $K$ is the batch size, $u(\cdot)$ is the uncertainty function that specifies the uncertainty of each sample, and $d(\cdot, \cdot)$ is a distance function that measures the distance between two samples. We adopt the max-min diversification objective, \ie, maximize the minimum uncertainty and distance of the selected set. The set selection function $f$ is defined as:

\begin{equation}
    f(\mathcal{B}_t) = \min_{x \in \mathcal{B}_t }u(x) + \min_{x, y\in\mathcal{B}_t}d(x,y).
\label{eq:set_sel}
\end{equation}

To solve \cref{eq:obj} via a 2-approximation greedy algorithm, we follow \cite{gollapudi2009axiomatic} to define a new distance function that combines the unary term with the pair-wise term: 
\begin{equation}
    d^{\prime}(x,y)= \frac{1}{2}(u(x)+u(y))+d(x,y).
\end{equation}
With $ d^{\prime}(\cdot)$, we can now solve \cref{eq:obj} efficiently using the algorithm summarized in \cref{alg:max-min}.

\begin{algorithm}
\caption{Active Selection via Max-Min Diversification}
\label{alg:max-min}
\begin{algorithmic}[1]
\Require Initial labeled set $\mathcal{L}_0$, Initial unlabelled set $\mathcal{U}_0$, Batch-size $K$, Maximum number of batches $T$
\Ensure Labeled set $\mathcal{L}_T$
\State For any $x \in \mathcal{U}_t$, define $d(x, \mathcal{B}_t\cup\mathcal{L}_t)=\min_{y\in \mathcal{B}_t\cup \mathcal{L}_t}d^{\prime}(x,y)$
\State $t = 0$
\While{$t < T$}
    \State Train Semantic-NeRF on $\mathcal{L}_t$
    \State $\mathcal{B}_t = \emptyset$
    \While{$|\mathcal{B}_t| < K$}
        \State $\hat{x} = \arg\max_{x \in \mathcal{U}_t} d(x,\mathcal{B}_t\cup\mathcal{L}_t)$
        \State $\mathcal{B}_t = \mathcal{B}_t \cup \{\hat{x}\}$
        \State $\mathcal{U}_t = \mathcal{U}_t \setminus \{\hat{x}\}$
    \EndWhile
    \State $\mathcal{L}_{t+1} = \mathcal{L}_t \cup \mathcal{B}_t$
    \State $\mathcal{U}_{t+1} = \mathcal{U}_t$
    \State $t = t + 1$
\EndWhile
\end{algorithmic}
\end{algorithm}

\subsection{Active Selection with 3D Geometric Constraint}

The set selection function \cref{eq:set_sel} requires the definition of uncertainty function $u(\cdot)$ and distance function $d(\cdot, \cdot)$. We use entropy for uncertainty estimation, which is computed as:
\begin{equation}
u(x) = -\sum_{c \in \set{C} } p_c(x)\log(p_c(x)),
\label{eq:ent}
\end{equation}
where $p_c(x)$ denotes the predicted probability for class $c$. For distance function, previous work \cite{sener2017active} represents each sample by a feature vector and computes the distance in the feature space, \ie, \(d_f(x, y) = \|F(x) - F(y)\|_2\), where \(F(\cdot)\) represent a feature extractor. However, the effectiveness of feature diversity relies on a good feature extractor, which is not always available due to the cold-start problem \cite{ji2023randomness} of AL (when the target model is used as feature extractor) or the domain gap (when a pretrained model is used as feature extractor). On the other hand, motivated by the observation that regions far away from each other in the spatial domain typically belong to different objects and have different semantics and appearance, we propose to enforce diversity in the 3D spatial domain to complement feature diversity.

More specifically, we take advantage of the property that Semantic-NeRF has learned to reconstruct the geometric of the scene and obtain the depth of each pixel via volume rendering \cite{nguyen2024semantically}:
\begin{equation}
D(\mathbf{r}) = \sum_{i=1}^N T_i(1-\exp(-\sigma_i\delta_i))t_i,
\end{equation}
where $\mathbf{r}$ is the ray passing the pixel, $\sigma_i$ is volume density, $T_i = \exp(-\sum_{j=1}^{i-1}\sigma_j\delta_j)$, and $\delta_i = t_{i+1}-t_i$ is the distance between adjacent points.

With the depth information, we can obtain the corresponding 3D coordinate for each pixel. The spatial diversity term is defined as the \(L_2\) distance between two samples \(x\) and \(y\), \ie, \(d_s(x, y) = \|C(x) - C(y)\|_2\), where \(C(\cdot)\) represents the averaged 3D coordinates of all pixels within the sample. 

Our distance function is then defined as:
\begin{equation}
d(x, y) = d_f(x, y) + d_s(x, y),
\end{equation}
where both distance terms are normalized to \([0, 1]\) before summation. Note that $d(x, y)$ is a metric as the summation of two metrics is still a metric.

\section{Experiments}
\label{sec:experiments}
In this section, we first provide the information for the datasets used in our experiments and the implementation details. We then investigate the effect of selection strategies by comparing our method with state-of-the-art methods, and the effect of selection granularity by performing active selection at both image and region level. Finally, we present ablation studies on design components and computational complexity analysis of our method. 

\subsection{Datasets}
\textbf{Replica}\cite{straub2019replicadatasetdigitalreplica} dataset comprises 18 different indoor scenes, including apartments, offices, and rooms. Each scene is captured with high-resolution RGB-D sensors and reconstructed using state-of-the-art techniques to ensure accuracy in geometry and texture. Following the setup in Semantic-NeRF, we sample every 5th frame from each scene sequence to form the training set, and use the frame in the middle of every two training frames as the test set.

\textbf{ScanNet}\cite{dai2017scannetrichlyannotated3dreconstructions} dataset includes over 1,500 indoor scenes captured using RGB-D sensors. These scenes encompass twenty different types, including Bedroom/Hotel, Living Room/Lounge, Bathroom, \etc. Similar to Replica, we uniformly sample frames from each scene sequence for the training set and use the intermediate frames as the test set.

\subsection{Implementation Details}

\textbf{Fully supervised training details}
We first establish an upper bound for all AL methods by training the Semantic-NeRF model using the full labelled training set. Following the setup of Semantic-NeRF, the training image is resized to $320\times 240$ and the learning rate is \( 5 \times 10^{-4} \). The model is trained for 100,000 iterations using Adam optimizer \cite{kingma2014adam}, where at each iteration, 1024 rays are randomly sampled from one image for loss computation. During the testing phase, the mean Intersection Over Union (mIoU) between the predicted and ground truth segmentation map is used as the evaluation metric.

\textbf{Batch training details}
We conduct experiments over 4 batches, starting from batch 0. The initial labelled pool of batch 0 is constructed by randomly selecting 5\% of regions. In the subsequent batch, we select additionally 5\% of regions to label using various active learning methods. After selection and annotation at each batch, the Semantic-NeRF model is re-trained from scratch using all the labelled data. The training parameters used are identical to those for fully supervised training.

\subsection{Effect of Selection Strategies}

In this section, we investigate the effect of selection strategies by comparing our method with the following methods:
\begin{itemize}
    \item \textbf{Random} This method selects samples randomly.

    \item \textbf{Entropy}\cite{shannon1948mathematical} This is an uncertainty-based method that selects samples with the highest entropy. 

     \item \textbf{CoreSet}\cite{sener2017active} This is a diversity-based method that uses the k-Center greedy algorithm to select samples that form a core-set of the training distribution. 
     
    \item \textbf{ViewAL}\cite{siddiqui2020viewal} This is a hybrid method that first selects samples with high viewpoint entropy and then selects sample that looks most different from other views. 
   
\end{itemize}

We follow ViewAL to divide image into irregularly-shaped regions, \ie, superpixels, and perform selection at superpixel level. Each image is divided into 300 non-overlapping superpixels using the SEEDS algorithm \cite{bergh2013seedssuperpixelsextractedenergydriven}. The entropy of a sample is calculated as the average entropy of all pixels within the sample. For feature representation in CoreSet and our method, we use the logits accumulated via volume rendering to represent each pixel, and the feature of a superpixel is computed by averaging the feature vectors of all pixels within it.

\begin{figure}[ht]
\vspace{-0.5cm}
\centering
\begin{subfigure}{0.22\textwidth}
	\includegraphics[width=\linewidth]{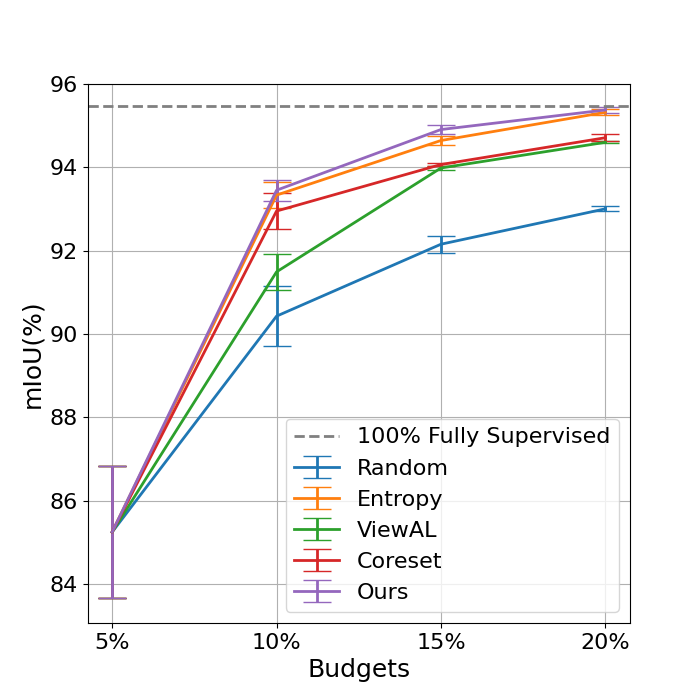}
	\caption{Scene0006 (Bedroom)}
	\label{fig:scene0006}
\end{subfigure}
\begin{subfigure}{0.22\textwidth}
	\includegraphics[width=\linewidth]{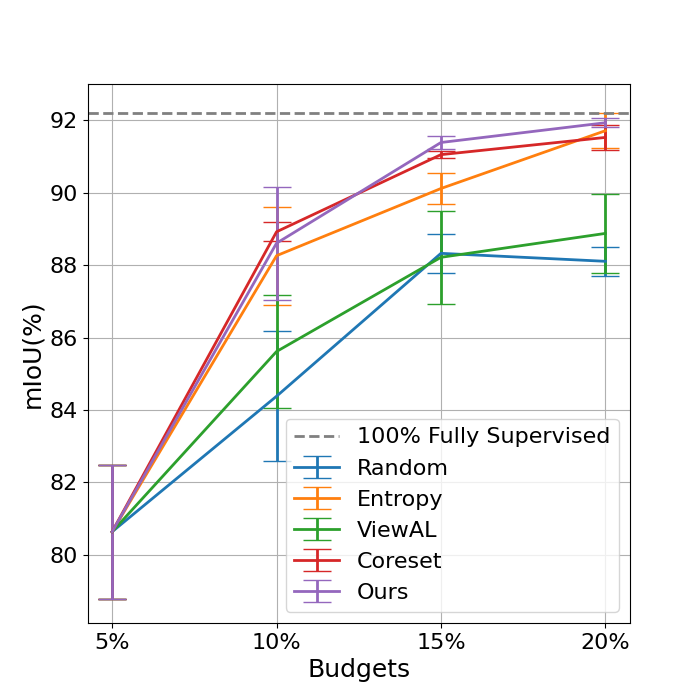}
	\caption{Scene0030 (Classroom)}
	\label{fig:scene0030}
\end{subfigure}

\caption{Active learning results on the ScanNet dataset. (a) Results for Scene0006 (Bedroom/Hotel); (b) Results for Scene0030 (Classroom). We plot the mean of three runs and the error bar indicates the standard deviation.}
\label{al_scannet}

\end{figure}

The benchmarking results on the ScanNet dataset is presented in \cref{al_scannet}, with \cref{fig:scene0006} depicting the results for Scene0006 (Bedroom/Hotel) and \cref{fig:scene0030} for Scene0030 (Classroom) (results on more scenes and qualitative results are provided in the supplementary). We observe that all AL methods can outperform Random, highlighting the effectiveness of AL in reducing the annotation cost for training Semantic-NeRF. ViewAL outperforms Random but lags behind the other methods, suggesting that viewpoint consistency is not an effective selection criterion for Semantic-NeRF. Entropy outperforms CoreSet for Scene0006, but the order is reversed for Scene0030. Our method is able to consistently outperform single-criterion-based method like Entropy and CoreSet, demonstrating the advantage of considering both uncertainty and diversity in selection. 

\begin{figure}[ht]
\vspace{-0.4cm}
\centering
\begin{subfigure}{0.22\textwidth}
	\includegraphics[width=\linewidth]{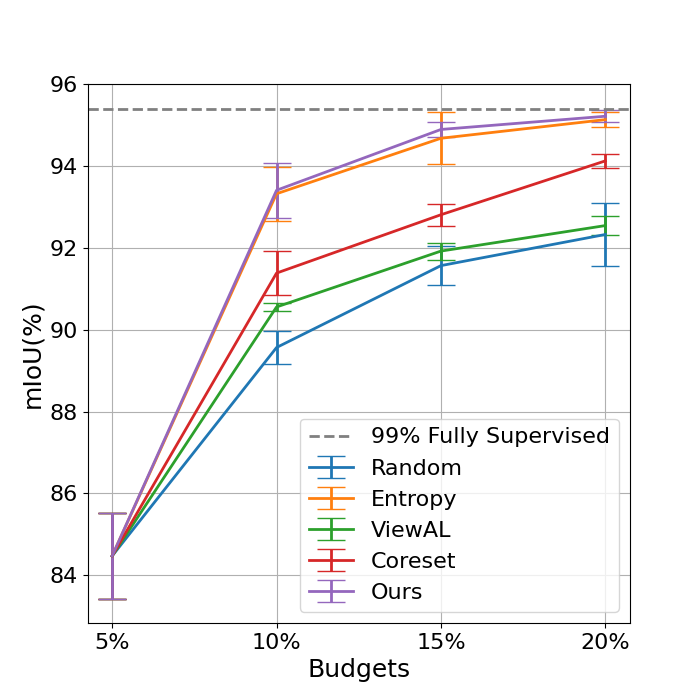}
	\caption{Room0}
	\label{fig:room0}
\end{subfigure}
\begin{subfigure}{0.22\textwidth}
	\includegraphics[width=\linewidth]{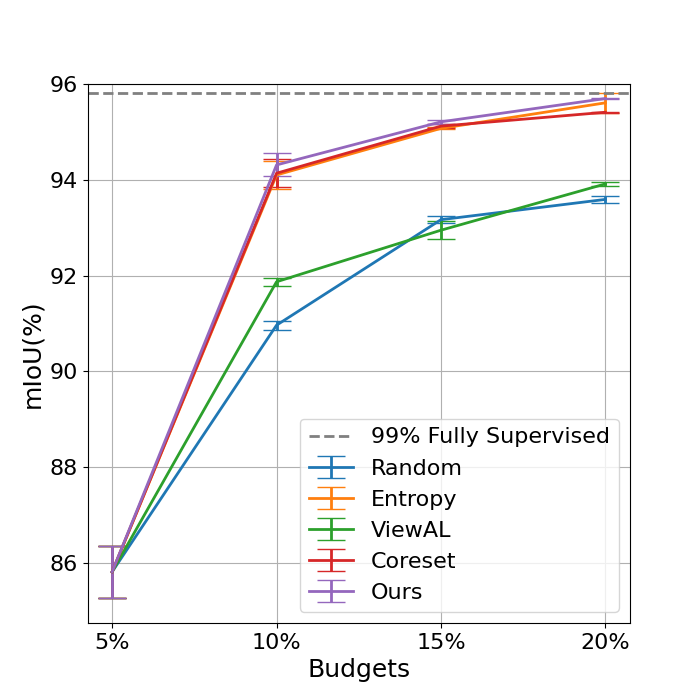}
	\caption{Office0}
	\label{fig:office0}
\end{subfigure}
\caption{Active learning results on the Replica dataset. (a) Results for Room0; (b) Results for Office0. We plot the mean of three runs and the error bar indicates the standard deviation.}
\label{al_replica}
\end{figure}

The benchmarking results on the Replica dataset is presented in \cref{al_replica}, with \cref{fig:room0} for scene Room0 and \cref{fig:office0} for scene Office0. We observe similar trend as for ScanNet, where all AL methods can outperform Random. However, the gain of ViewAL over Random is marginal, reiterating the ineffectiveness of viewpoint consistency for Semantic-NeRF. For Room0 (\cref{fig:room0}), Entropy significantly outperforms CoreSet, while for Office0 ( \cref{fig:office0}), the two perform comparably. Our method, being a hybrid strategy, consistently outperforms all the competing methods under different budgets.

We further look into the amount of reduction in annotation that AL brings compared to Random. We report the amount of annotation required for various method to achieve the same performance of Random at 20\% budget in \cref{tab:budgets}. Our method achieves more than 2$\times$ reduction in annotation cost compared to Random sampling. 
\begin{table}[ht]

\caption{Amount of annotation required for various methods to achieve the same performance. Our method achieves more than 2$\times$ reduction in annotation cost compared to Random.}
\centering

\begin{tabular}{crrrr}
\toprule
Method   & Scene0006       & Scene0030      & Room0 & Office0\\
\midrule
Random & 20.00\%  & 20.00\% & 20.00\% & 20.00\% \\
ViewAL & 12.69\% & 14.58\%  & 18.04\% & 18.10\%  \\
Entropy & 9.56\% & 9.62\%   & 9.56\% & 9.43\% \\
CoreSet & 9.83\% & \textbf{9.13}\%  & 13.24\% & 9.41\% \\
Ours & \textbf{9.50\%} & 9.44\%   & \textbf{9.28\%} & \textbf{9.34\%} \\
\bottomrule
\end{tabular}
%}
\label{tab:budgets}
\vspace{-0.2cm}
\end{table}

\subsection{Effect of Selection Granularity}
We investigate the effect of selection granularity by performing selection at both image and superpixel level for three methods, namely, Random, Entropy, and CoreSet. The results are reported in \cref{tab:ablation_study1}. We observe that both Entropy and CoreSet perform better at superpixel level, exhibiting smaller standard deviations and thus greater stability. These results suggest that superpixel-level selection is more cost-effective for AL of Semantic-NeRF.

\begin{table}[!]

\caption{Effect of selection granularity. Both Entropy and CoreSet perform better at superpixel level, suggesting that superpixel-level selection is more cost-effective for AL of Semantic-NeRF. We report the mean and standard deviation of three runs on scene Room0 of the Replica dataset.}
\centering

\begin{tabular}{cccccc}
\toprule
Method           & Batch 0           & Batch 1      & Batch 2 & Batch 3\\
\midrule
\multicolumn{5}{l}{Random} \\
\midrule
Image & \textbf{84.95(2.58)}  & 87.58 (4.74) & \textbf{92.18 (1.62)} & 92.27 (0.65) \\
Superpixel  & 84.75(1.04) & \textbf{89.57 (0.41)} & 91.57 (0.47) & \textbf{92.33 (0.78)}  \\
\midrule
\multicolumn{5}{l}{Entropy} \\
\midrule
Image & \textbf{84.95(2.58)}  & 88.22 (1.40) & 90.68 (0.30) & 92.15 (0.76) \\
Superpixel & 84.75(1.04) & \textbf{93.12 (0.64)} & \textbf{94.59 (0.71)} & \textbf{95.18 (0.19)} \\
\midrule
\multicolumn{5}{l}{CoreSet} \\
\midrule
Image & \textbf{84.95(2.58)}  & 90.46 (0.69) & 91.43 (0.23) & 93.02 (0.56) \\
Superpixel & 84.75(1.04)  & \textbf{91.56 (0.52)} & \textbf{92.84 (0.31)} & \textbf{94.07 (0.18)} \\
\bottomrule
\end{tabular}
%}
\label{tab:ablation_study1}
%\vspace{-0.1cm}
\end{table}

\subsection{Ablation Studies}
In this section, we investigate the effect of the three terms in the distance function, \ie, entropy, feature diversity and spatial diversity, on model performance. We experiment with different combination of the terms, resulting in five variants, namely, Entropy, Feature (\ie, CoreSet), Entropy+Feature, Entropy+Spatial and Entropy+Feature+Spatial (\ie, our method). The results are reported in \cref{tab:ablation_study2}. We observe that adding the proposed spatial diversity term can effectively improve the performance of Entropy and Entropy+Feature, while the feature diversity term alone cannot achieve this effect. This demonstrates the effectiveness of the proposed 3D geometric constraint in selecting informative samples.

\begin{table}[ht]
\caption{Ablation studies on the three terms in distance function. The proposed spatial diversity term can effectively improve the performance of Entropy and Entropy+Feature, while the feature diversity term alone cannot achieve this effect. We report the mean and standard deviation of three runs on scene Room0 of the Replica dataset.}
\centering
\begin{tabular}{ccccccc}
\toprule
Entropy          & Feature          & Spatial         & Batch 1      & Batch 2 & Batch 3\\
\midrule
\checkmark &             &             & 93.12 (0.64) & 94.59 (0.71) & \textbf{95.18 (0.19)} \\
           & \checkmark  &             & 91.56 (0.52) & 92.84 (0.31) & 94.07 (0.18) \\
\checkmark & \checkmark  &             & 93.00 (0.63) & 94.56 (0.60) & 95.14 (0.07) \\
\checkmark &             & \checkmark  & 93.20 (0.86) & 94.93 (0.20) & 95.15 (0.18) \\
\checkmark & \checkmark  & \checkmark  & \textbf{93.32 (0.76)} & \textbf{94.95 (0.18)} & 95.16 (0.14) \\
\bottomrule
\end{tabular}
\label{tab:ablation_study2}
\vspace{-0.2cm}
\end{table}

\subsection{Computational Complexity Analysis}
We analyze the computational complexity of the proposed selection algorithm as below. Let $n_b$ denote the number of selected samples, $n_u$ the total number of unlabelled samples, and $f_{dim}$ the input sample dimension. Due to greedy selection, we can avoid quadratic complexity and  the time complexity of our selection algorithm is $O(n_b\cdot n_u \cdot f_{dim})$. This complexity is similar to the k-Center greedy algorithm used in CoreSet \cite{sener2017active}. However, our method considers both uncertainty and diversity in selection, while CoreSet only considers diversity.

\section{Conclusions}
\label{sec:conclusion}
In this work, we perform a comprehensive study on active learning for semantically-aware NeRF. We experiment with various design choices, including image-level \vs region-level selection, uncertainty-based, diversity-based and hybrid selection strategies. Motivated by the limitation of feature diversity, we propose to take into account 3D geometric constraint and enforce diversity in the 3D spatial domain. We incorporate the geometric constraint into the result diversification framework and solve it efficiently using a 2-approximation greedy algorithm. We evaluate the effectiveness of the proposed method on Replica and ScanNet datasets. Experimental results demonstrate that our method consistently outperforms competing methods under various annotation budgets. Our work demonstrates that active learning can effectively reduce the annotation cost for training semantically-aware NeRF, achieving more than 2$\times$ reduction in annotation cost compared to random sampling, and thus serves as a promising solution for label-efficient training of semantically-aware NeRF.

\subsubsection*{Acknowledgments}
This work is supported by the Agency for Science, Technology and Research (A*STAR) under its MTC Programmatic Funds (Grant No. M23L7b0021).

\bibliographystyle{IEEEbib}
\bibliography{sec/references}

\begin{thebibliography}{10}

\bibitem{mildenhall2021nerf}
Ben Mildenhall, Pratul~P Srinivasan, Matthew Tancik, Jonathan~T Barron, Ravi Ramamoorthi, and Ren Ng,
\newblock ``Nerf: Representing scenes as neural radiance fields for view synthesis,''
\newblock {\em Communications of the ACM}, vol. 65, no. 1, pp. 99--106, 2021.

\bibitem{gao2022nerf}
Kyle Gao, Yina Gao, Hongjie He, Dening Lu, Linlin Xu, and Jonathan Li,
\newblock ``Nerf: Neural radiance field in 3d vision, a comprehensive review,''
\newblock {\em arXiv preprint arXiv:2210.00379}, 2022.

\bibitem{zhang2020nerf++}
Kai Zhang, Gernot Riegler, Noah Snavely, and Vladlen Koltun,
\newblock ``Nerf++: Analyzing and improving neural radiance fields,''
\newblock {\em arXiv preprint arXiv:2010.07492}, 2020.

\bibitem{yu2021pixelnerf}
Alex Yu, Vickie Ye, Matthew Tancik, and Angjoo Kanazawa,
\newblock ``pixelnerf: Neural radiance fields from one or few images,''
\newblock in {\em Proceedings of the IEEE/CVF conference on computer vision and pattern recognition}, 2021, pp. 4578--4587.

\bibitem{zhi2021place}
Shuaifeng Zhi, Tristan Laidlow, Stefan Leutenegger, and Andrew~J Davison,
\newblock ``In-place scene labelling and understanding with implicit scene representation,''
\newblock in {\em Proceedings of the IEEE/CVF International Conference on Computer Vision}, 2021, pp. 15838--15847.

\bibitem{ren2021survey}
Pengzhen Ren, Yun Xiao, Xiaojun Chang, Po-Yao Huang, Zhihui Li, Brij~B Gupta, Xiaojiang Chen, and Xin Wang,
\newblock ``A survey of deep active learning,''
\newblock {\em ACM computing surveys (CSUR)}, vol. 54, no. 9, pp. 1--40, 2021.

\bibitem{siddiqui2020viewal}
Yawar Siddiqui, Julien Valentin, and Matthias Nie{\ss}ner,
\newblock ``Viewal: Active learning with viewpoint entropy for semantic segmentation,''
\newblock in {\em Proceedings of the IEEE/CVF conference on computer vision and pattern recognition}, 2020, pp. 9433--9443.

\bibitem{chen2018encoder}
Liang-Chieh Chen, Yukun Zhu, George Papandreou, Florian Schroff, and Hartwig Adam,
\newblock ``Encoder-decoder with atrous separable convolution for semantic image segmentation,''
\newblock in {\em Proceedings of the European conference on computer vision (ECCV)}, 2018, pp. 801--818.

\bibitem{liu2023unsupervised}
Zhizheng Liu, Francesco Milano, Jonas Frey, Roland Siegwart, Hermann Blum, and Cesar Cadena,
\newblock ``Unsupervised continual semantic adaptation through neural rendering,''
\newblock in {\em Proceedings of the IEEE/CVF Conference on Computer Vision and Pattern Recognition}, 2023, pp. 3031--3040.

\bibitem{fu2022panoptic}
Xiao Fu, Shangzhan Zhang, Tianrun Chen, Yichong Lu, Lanyun Zhu, Xiaowei Zhou, Andreas Geiger, and Yiyi Liao,
\newblock ``Panoptic nerf: 3d-to-2d label transfer for panoptic urban scene segmentation,''
\newblock in {\em 2022 International Conference on 3D Vision (3DV)}. IEEE, 2022, pp. 1--11.

\bibitem{liu2023weakly}
Kunhao Liu, Fangneng Zhan, Jiahui Zhang, Muyu Xu, Yingchen Yu, Abdulmotaleb El~Saddik, Christian Theobalt, Eric Xing, and Shijian Lu,
\newblock ``Weakly supervised 3d open-vocabulary segmentation,''
\newblock {\em Advances in Neural Information Processing Systems}, vol. 36, pp. 53433--53456, 2023.

\bibitem{zhi2021ilabel}
Shuaifeng Zhi, Edgar Sucar, Andre Mouton, Iain Haughton, Tristan Laidlow, and Andrew~J Davison,
\newblock ``ilabel: Interactive neural scene labelling,''
\newblock {\em arXiv preprint arXiv:2111.14637}, 2021.

\bibitem{tang2023scene}
Songlin Tang, Wenjie Pei, Xin Tao, Tanghui Jia, Guangming Lu, and Yu-Wing Tai,
\newblock ``Scene-generalizable interactive segmentation of radiance fields,''
\newblock in {\em Proceedings of the 31st ACM International Conference on Multimedia}, 2023, pp. 6744--6755.

\bibitem{beluch2018power}
William~H Beluch, Tim Genewein, Andreas N{\"u}rnberger, and Jan~M K{\"o}hler,
\newblock ``The power of ensembles for active learning in image classification,''
\newblock in {\em Proceedings of the IEEE conference on computer vision and pattern recognition}, 2018, pp. 9368--9377.

\bibitem{sener2017active}
Ozan Sener and Silvio Savarese,
\newblock ``Active learning for convolutional neural networks: A core-set approach,''
\newblock {\em arXiv preprint arXiv:1708.00489}, 2017.

\bibitem{ash2019deep}
Jordan~T Ash, Chicheng Zhang, Akshay Krishnamurthy, John Langford, and Alekh Agarwal,
\newblock ``Deep batch active learning by diverse, uncertain gradient lower bounds,''
\newblock {\em arXiv preprint arXiv:1906.03671}, 2019.

\bibitem{he2024hybrid}
Yinan He, Lile Cai, Jingyi Liao, and Chuan-Sheng Foo,
\newblock ``Hybrid active learning with uncertainty-weighted embeddings,''
\newblock {\em Transactions on Machine Learning Research}, 2024.

\bibitem{wen2024active}
Ziting Wen, Oscar Pizarro, and Stefan Williams,
\newblock ``Active self-semi-supervised learning for few labeled samples,''
\newblock {\em Neurocomputing}, p. 128772, 2024.

\bibitem{casanova2020reinforced}
Arantxa Casanova, Pedro~O Pinheiro, Negar Rostamzadeh, and Christopher~J Pal,
\newblock ``Reinforced active learning for image segmentation,''
\newblock {\em arXiv preprint arXiv:2002.06583}, 2020.

\bibitem{cai2021revisiting}
Lile Cai, Xun Xu, Jun~Hao Liew, and Chuan~Sheng Foo,
\newblock ``Revisiting superpixels for active learning in semantic segmentation with realistic annotation costs,''
\newblock in {\em Proceedings of the IEEE/CVF conference on computer vision and pattern recognition}, 2021, pp. 10988--10997.

\bibitem{cai2021exploring}
Lile Cai, Xun Xu, Lining Zhang, and Chuan-Sheng Foo,
\newblock ``Exploring spatial diversity for region-based active learning,''
\newblock {\em IEEE Transactions on Image Processing}, vol. 30, pp. 8702--8712, 2021.

\bibitem{wu2022entropy}
Jiaxi Wu, Jiaxin Chen, and Di~Huang,
\newblock ``Entropy-based active learning for object detection with progressive diversity constraint,''
\newblock in {\em Proceedings of the IEEE/CVF Conference on Computer Vision and Pattern Recognition}, 2022, pp. 9397--9406.

\bibitem{lyu2023box}
Mengyao Lyu, Jundong Zhou, Hui Chen, Yijie Huang, Dongdong Yu, Yaqian Li, Yandong Guo, Yuchen Guo, Liuyu Xiang, and Guiguang Ding,
\newblock ``Box-level active detection,''
\newblock in {\em Proceedings of the IEEE/CVF Conference on Computer Vision and Pattern Recognition}, 2023, pp. 23766--23775.

\bibitem{liao2024box}
Jingyi Liao, Xun Xu, Chuan-Sheng Foo, and Lile Cai,
\newblock ``Box-level class-balanced sampling for active object detection,''
\newblock in {\em 2024 IEEE International Conference on Image Processing (ICIP)}. IEEE, 2024, pp. 701--707.

\bibitem{zhang2024employing}
Licheng Zhang, Siew-Kei Lam, Dingsheng Luo, and Xihong Wu,
\newblock ``Employing feature mixture for active learning of object detection,''
\newblock {\em Neurocomputing}, vol. 594, pp. 127883, 2024.

\bibitem{yoo2019learning}
Donggeun Yoo and In~So Kweon,
\newblock ``Learning loss for active learning,''
\newblock in {\em Proceedings of the IEEE/CVF conference on computer vision and pattern recognition}, 2019, pp. 93--102.

\bibitem{liu2021influence}
Zhuoming Liu, Hao Ding, Huaping Zhong, Weijia Li, Jifeng Dai, and Conghui He,
\newblock ``Influence selection for active learning,''
\newblock in {\em Proceedings of the IEEE/CVF international conference on computer vision}, 2021, pp. 9274--9283.

\bibitem{wu2021hal}
Xing Wu, Cheng Chen, Mingyu Zhong, and Jianjia Wang,
\newblock ``Hal: Hybrid active learning for efficient labeling in medical domain,''
\newblock {\em Neurocomputing}, vol. 456, pp. 563--572, 2021.

\bibitem{shannon1948mathematical}
Claude~Elwood Shannon,
\newblock ``A mathematical theory of communication,''
\newblock {\em The Bell system technical journal}, vol. 27, no. 3, pp. 379--423, 1948.

\bibitem{pan2022activenerf}
Xuran Pan, Zihang Lai, Shiji Song, and Gao Huang,
\newblock ``Activenerf: Learning where to see with uncertainty estimation,''
\newblock in {\em European Conference on Computer Vision}. Springer, 2022, pp. 230--246.

\bibitem{gollapudi2009axiomatic}
Sreenivas Gollapudi and Aneesh Sharma,
\newblock ``An axiomatic approach for result diversification,''
\newblock in {\em Proceedings of the 18th international conference on World wide web}, 2009, pp. 381--390.

\bibitem{ji2023randomness}
Yilin Ji, Daniel Kaestner, Oliver Wirth, and Christian Wressnegger,
\newblock ``Randomness is the root of all evil: more reliable evaluation of deep active learning,''
\newblock in {\em Proceedings of the IEEE/CVF Winter Conference on Applications of Computer Vision}, 2023, pp. 3943--3952.

\bibitem{nguyen2024semantically}
Thang-Anh-Quan Nguyen, Amine Bourki, M{\'a}ty{\'a}s Macudzinski, Anthony Brunel, and Mohammed Bennamoun,
\newblock ``Semantically-aware neural radiance fields for visual scene understanding: A comprehensive review,''
\newblock {\em arXiv preprint arXiv:2402.11141}, 2024.

\bibitem{straub2019replicadatasetdigitalreplica}
Julian Straub, Thomas Whelan, Lingni Ma, Yufan Chen, Erik Wijmans, Simon Green, Jakob~J. Engel, Raul Mur-Artal, Carl Ren, Shobhit Verma, Anton Clarkson, Mingfei Yan, Brian Budge, Yajie Yan, Xiaqing Pan, June Yon, Yuyang Zou, Kimberly Leon, Nigel Carter, Jesus Briales, Tyler Gillingham, Elias Mueggler, Luis Pesqueira, Manolis Savva, Dhruv Batra, Hauke~M. Strasdat, Renzo~De Nardi, Michael Goesele, Steven Lovegrove, and Richard Newcombe,
\newblock ``The replica dataset: A digital replica of indoor spaces,'' 2019.

\bibitem{dai2017scannetrichlyannotated3dreconstructions}
Angela Dai, Angel~X. Chang, Manolis Savva, Maciej Halber, Thomas Funkhouser, and Matthias Nießner,
\newblock ``Scannet: Richly-annotated 3d reconstructions of indoor scenes,'' 2017.

\bibitem{kingma2014adam}
Diederik~P Kingma and Jimmy Ba,
\newblock ``Adam: A method for stochastic optimization,''
\newblock {\em arXiv preprint arXiv:1412.6980}, 2014.

\bibitem{bergh2013seedssuperpixelsextractedenergydriven}
Michael~Van den Bergh, Xavier Boix, Gemma Roig, and Luc~Van Gool,
\newblock ``Seeds: Superpixels extracted via energy-driven sampling,'' 2013.

\end{thebibliography}

\appendix
\section{Appendix}
\subsection{Qualitative Results}
\begin{figure*}[!]
\centering
\includegraphics[width=0.95\linewidth]{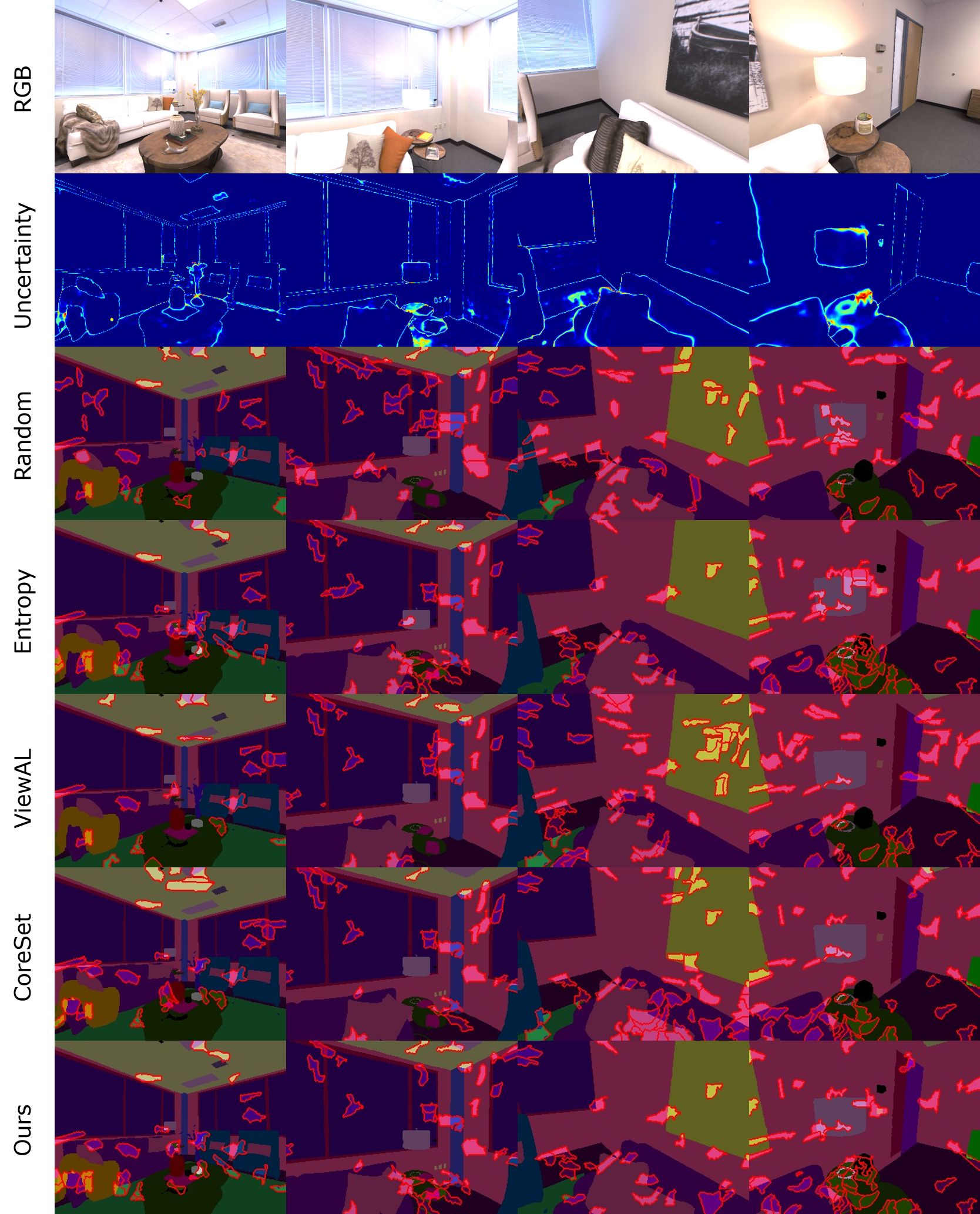}
\caption{Visualization of regions selected by different methods in the first batch for scene Room0 of the Replica dataset. The selected superpixels are highlighted. The second row displays the uncertainty map estimated by current model. Our method allows the annotation budget to be spent on more informative and diverse regions.}
\label{fig:visualization}
\end{figure*}

We visualize the regions selected by various methods for scene Room0 of the Replica dataset in \cref{fig:visualization}. In the second row of \cref{fig:visualization}, we display the uncertainty map estimated by current model for each image. We notice that regions of high uncertainty typically correspond to small objects (\eg, side tables, items on the table) and object boundaries. Entropy selects samples from the most uncertain regions, but the selected samples tend to be clustered together and may be redundant. Compared to Entropy, our method avoids selecting too many neighboring regions and allows the annotation budget to be spent on more diverse regions. Compared to CoreSet, our method avoids selecting regions in uninformative regions. It can also been observed that ViewAL is not effective in selecting informative regions for Semantic-NeRF, wasting much annotation budget for low-uncertainty regions on wall and painting (\eg, third column of \cref{fig:visualization}).

\subsection{Results on additional scenes}
\begin{figure*}[ht]
\centering
\vspace{-0.0cm}
\begin{subfigure}{0.45\textwidth}
	\includegraphics[width=\linewidth]{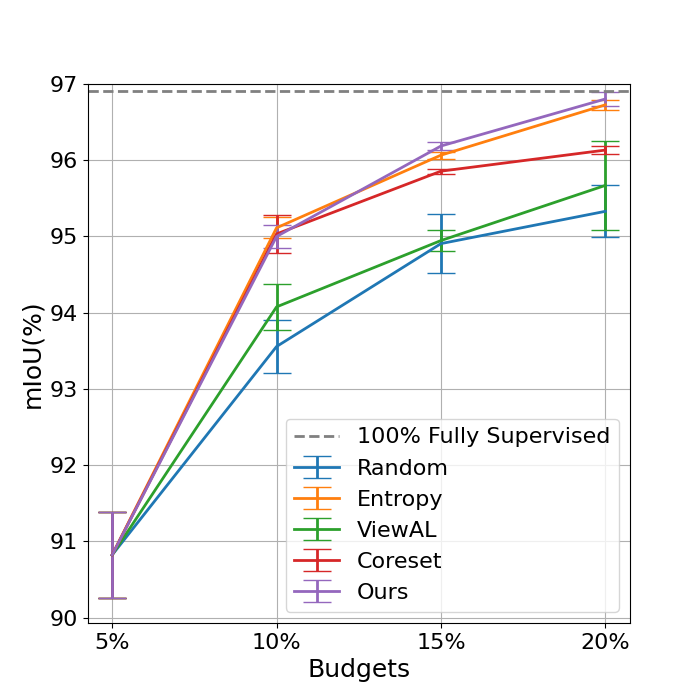}
	\caption{Scene0005 (Misc.)}
	\label{fig:scene0005}
\end{subfigure}
\begin{subfigure}{0.45\textwidth}
	\includegraphics[width=\linewidth]{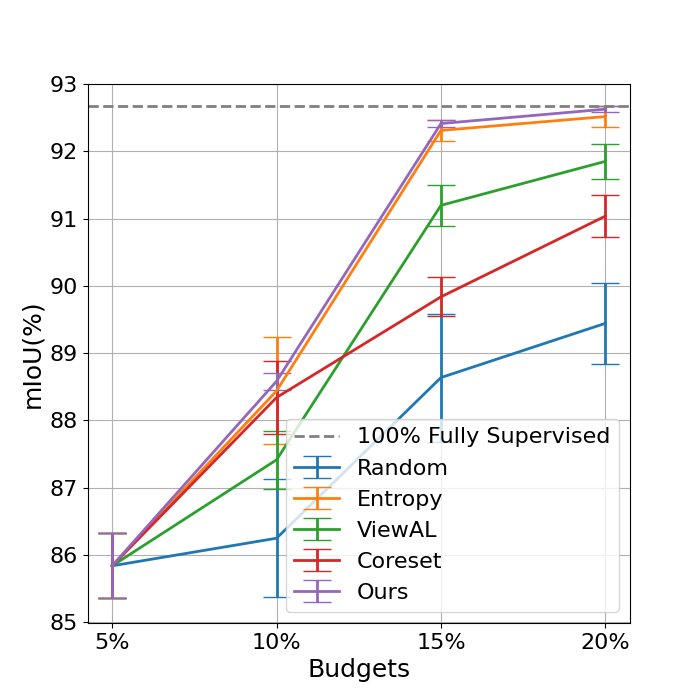}
	\caption{Scene0009 (Bathroom)}
	\label{fig:scene0009}
\end{subfigure}
\begin{subfigure}{0.45\textwidth}
	\includegraphics[width=\linewidth]{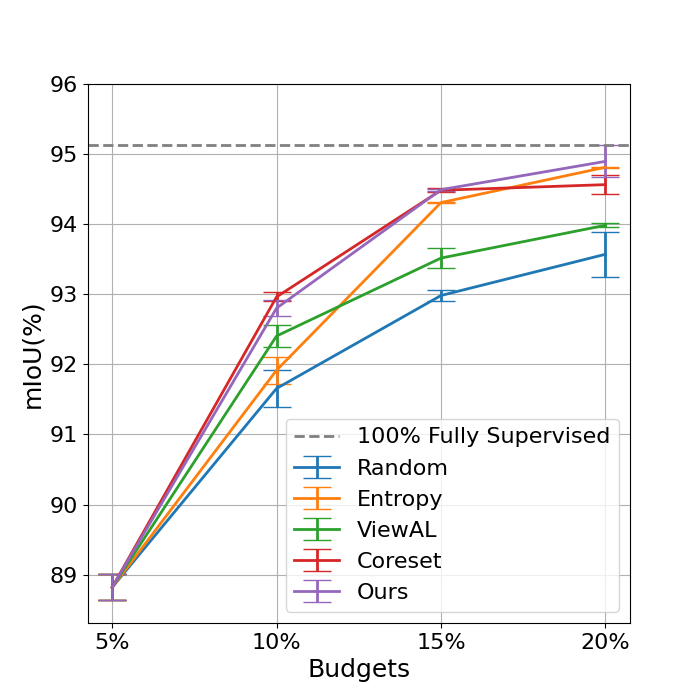}
	\caption{Scene0010 (Office)}
	\label{fig:scene0010}
\end{subfigure}
\begin{subfigure}{0.45\textwidth}
	\includegraphics[width=\linewidth]{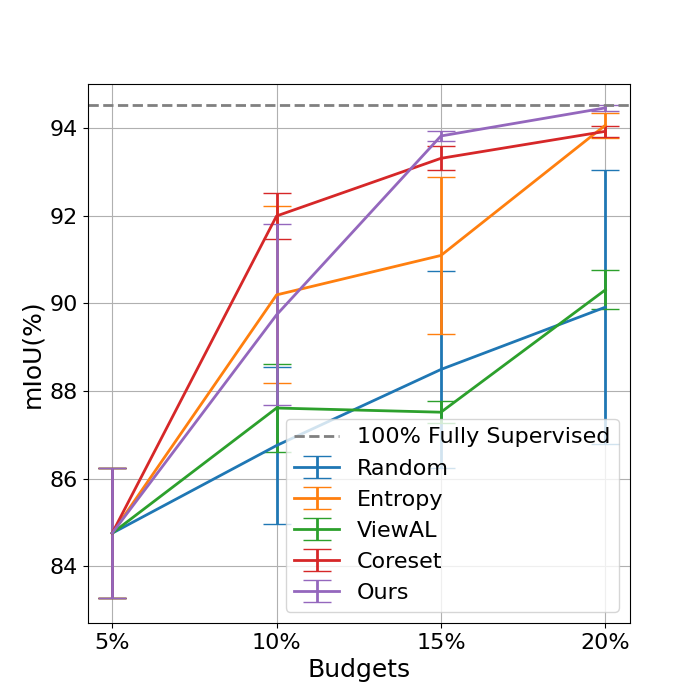}
	\caption{Scene0011 (Kitchen)}
	\label{fig:scene0011}
\end{subfigure}
\caption{Additional active learning results on the ScanNet dataset. (a) Results for Scene0005 (Misc.); (b) Results for Scene0009 (Bathroom); (c) Results for Scene0010 (Office); (d) Results for Scene0011 (Kitchen). We plot the mean of three runs and the error bar indicates the standard deviation.}
\label{fig:al_scannet_more}
\vspace{-0.0cm}
\end{figure*}

We provide additional results on scenes that are sampled from different categories to demonstrate the generalizability and robustness of the proposed method in \cref{fig:al_scannet_more}. We observe that the performance of different methods can vary for difference scenes, \eg, Entropy outperforms CoreSet for Scene0005 (Misc.) and Scene0009 (Bathroom), but the order is reversed for Scene0010 (Office) and Scene0011 (Kitchen); ViewAL performs marginally better than Random for Scene0005 (Misc.) and Scene0011 (Kitchen), while the improvement is more significant for Scene0009 (Bathroom) and Scene0010 (Office). Our method is able to consistently outperform other methods across different scenes, demonstrating the advantage of the proposed hybrid selection strategy in handling datasets of different characteristics.    
\end{document}